\documentclass[conference]{IEEEtran}

\usepackage{amsmath,amssymb,amsfonts}
\usepackage{algorithmic}
\usepackage{graphicx}
\usepackage{textcomp}
\usepackage{xcolor}
\usepackage{booktabs}
\usepackage{tabularx}
\usepackage{float}
\usepackage{stfloats}
\usepackage[colorlinks=true, citecolor=blue, linkcolor=blue, urlcolor=blue]{hyperref}
\usepackage[T1]{fontenc}
\usepackage{mathptmx}

\begin{document}

\title{Diversity-Based Active Learning: An Evaluation of Metric Spaces for Active Learning Selection}

\author{\IEEEauthorblockN{Siddharth Chilamkur}
\IEEEauthorblockA{\textit{University of California, Berkeley} \\
schilamkur@berkeley.edu}
\and
\IEEEauthorblockN{Dorit S. Hochbaum}
\IEEEauthorblockA{\textit{University of California, Berkeley} \\
dhochbaum@berkeley.edu}
}

\maketitle

\begin{abstract}
With rapid advancement over the last few years, many different methods are now widely used for classification. However, training these models requires substantial labeled data. Active Learning is a potential solution to this problem. Pool-based active learning minimizes costs by querying only the most informative samples from an unlabeled dataset. Diversity-based approaches, on the other hand, attempt to select a representative subset of the data. There are many different objectives for determining the selection process, including exact $K$-center, exact $K$-median, and Greedy $K$-center. In this paper, we will focus on evaluating the performance of Greedy $K$-center across a variety of metric spaces: the raw feature space, a Linear Discriminant Analysis (LDA) space, and a model-derived probability space (with and without entropy-based weighting). Using Random Forest classifiers as a baseline evaluator, our empirical results on synthetic and real-world datasets demonstrate that mapping unlabeled instances into a predictive probability space and weighting the result by entropy often dominates the other options for active learning selection with Greedy $K$-center.
\end{abstract}

\begin{IEEEkeywords}
Active Learning, Greedy $K$-center, Random Forest, Dimensionality Reduction, Classification
\end{IEEEkeywords}

\section{Introduction}

Supervised machine learning relies quite heavily on the availability of large datasets that are correctly labeled. In domains where acquiring these labels becomes prohibitively expensive (e.g. medical images, specialized financial datasets) or where conducting the physical experiments to generate ground-truth is too time-consuming, the manual data labeling process frequently becomes the primary bottleneck in model deployment. Active learning (AL) provides a mathematical framework to strategically query only the most informative data points from a large, static pool of unlabeled data. By doing so, AL allows models to achieve competitive predictive accuracy with a minimal budget of labels, directly reducing the associated time and financial costs.

Among the different AL frameworks, pool-based active learning is overwhelmingly common in modern applications due to its alignment with how data is naturally collected and stored. With a pool-based setup, an algorithm iteratively evaluates a large, static set of unlabeled instances, scoring them using a specific mathematical heuristic. The algorithm then selects a batch of the highest-scoring samples to be labeled, usually by a human oracle or an expensive secondary process. Once these specific samples are labeled, they are incorporated into the training set, the model is retrained on the updated data, and the cycle repeats until the desired amount of labeled data is reached or a performance threshold is met.

The heuristics used to evaluate and score the unlabeled samples traditionally rely heavily on measures of uncertainty or model disagreement. A foundational framework for identifying these informative samples is the Query-by-Committee (QBC) algorithm, which measures the disagreement among a committee of differing models to perform selective sampling \cite{freund1997selective}. Similarly, margin-based uncertainty sampling queries points closest to a support vector machine's decision boundary. However, methods that exclusively query near the decision boundary or strictly prioritize uncertainty can suffer heavily from sampling bias. They are prone to querying localized clusters of outliers, overlapping noise, or redundant samples, effectively ignoring large, unmapped regions of the underlying data distribution.

To deal with some of these issues, diversity-based methods look to select a subset of points that represent the global data distribution, making sure that the model explores all of the data domain evenly. These approaches utilize a variety of combinatorial objective functions that attempt to maximize the dispersion or the representative relationship between the selected facilities (the samples) and the remaining unlabeled pool. Common objective functions in this domain include minimizing the sum of distances (the $K$-median problem), minimizing the maximum distance to a center (the $K$-center problem), and maximizing the minimum distance between any pair of selected centers ($K$-max-min-dispersion).

Recent theoretical and empirical work by Bodine and Hochbaum \cite{bodine2026kcoverage} discusses these different objectives and evaluates their heuristic performance in these contexts. A key finding from their work, which heavily influences the direction of this paper, is that optimizing for the exact $K$-center is not only computationally expensive, but empirically seems to perform worse for active learning tasks compared to a Greedy approximation. They also show that optimizing for exact $K$-median provides theoretically (see (\ref{eq:kmedian})) a lower loss function and better performance empirically than that of the optimal $K$-center solution.
Specifically, using the notation $S$ for the set of queried points, $\ell(x_i, y_i; S)$ the loss function for sample $x_i$ with true label $y_i$ when trained on set $S$, $\delta_i = \min_{j \in S} ||x_i - x_j||$  the distance from a point to its closest selected center, $\lambda_l$ and $\lambda_\mu$ the Lipschitz constants, $L$ the loss bound, and $C$ the number of classes, it was proved in \cite{bodine2026kcoverage} that:
\begin{equation}
\begin{aligned}
    \frac{1}{n}\sum_{i\in[n]} \ell(x_i, y_i; S) &- \frac{1}{|S|}\sum_{j \in S} \ell(x_j, y_j; S) \\
    &\leq \left[ \frac{1}{n}\sum_{i\in[n]} \delta_i \right] \left( \lambda_l + \lambda_\mu L C \right) + \sqrt{\frac{L^2 \log(1/\gamma)}{2n}}
\end{aligned}
\label{eq:kmedian}
\end{equation}
The coefficient $\frac{1}{n}\sum_{i\in[n]} \delta_i$ in the right hand side is the $K$-median objective, which is always lower than the respective $K$-center objective, $\max_{i\in[n]}\delta_i$,
which was the coefficient in the bound proved in the Core-Set approach \cite{sener2017active} and used as justification for using the $K$-center objective.
Although exact $K$-median empirically dominates in performance, it is computationally too expensive and fast heuristics for $K$-median perform badly. In addition, the tests in \cite{bodine2026kcoverage} show that the Greedy $K$-center approach dominates the exact $K$-center formulation for the active learning set selection problem. Because of this, the main focus of this paper is not on attempting to solve the exact $K$-median or exact $K$-center objectives, but rather on evaluating the metric spaces in which it is necessary to consider aspects of the data beyond just geometric projections.

The exact $K$-center problem is NP-Hard. Hochbaum and Shmoys \cite{hochbaum1985best} provided a best possible 2-approximation algorithm (in this case, best possible refers to the fact that the 2-$\epsilon$ approximation to the $K$-center problem is NP-hard for any $\epsilon>0$). An alternative simple Greedy algorithm was devised by Gonzales, also achieving 2-approximation \cite{gonzalez1985clustering}. The Gonzalez algorithm, referred to here as the Greedy $K$-center algorithm, achieves this by iteratively selecting the unlabeled point whose distance to the closest previously selected cluster center is maximal.

The connection between active learning and Greedy $K$-center solution was shown under the name ``Core-Set approach" \cite{sener2017active}. The effectiveness of the Greedy $K$-center heuristic was shown there by applying it to the learned feature embeddings of Convolutional Neural Networks. While this Core-Set paper demonstrated the effectiveness of Greedy $K$-center using these feature embeddings, the literature at large is unspecific regarding the best generalized metric space for these calculations across different data types. In practice, the Greedy $K$-center approach has been shown to be a strong baseline for active learning that is difficult to consistently beat. However, as we show here, its success is determined by the space in which the distances are calculated. If the space does not accurately reflect the classification task, maximizing distance might simply maximize the selection of noise.

In high-dimensional spaces, standard physical distance metrics (e.g. $L_2$ norm) suffer from the curse of high dimensionality, becoming significantly less informative as the distance between any two random points begins to converge, \cite{aggarwal2001surprising}. To address this, dimensionality reduction is frequently applied before attempting any type of clustering or sampling. Principal Component Analysis (PCA) is a standard unsupervised technique used to project data into a lower-dimensional space that maximizes overall variance. However, because PCA does not incorporate knowledge about class labels, the directions of maximum variance often do not align with the directions that maximize class separability. This can lead to projecting distinct classes onto the same axis and collapsing them together in the new space, leading to suboptimal active learning selection in a PCA-reduced space.

On the other hand, Linear Discriminant Analysis (LDA) can outperform Principal Component Analysis (PCA) in certain classification-focused tasks because it maximizes between-class separability rather than simply total variance. A classic paper providing a direct comparative analysis where LDA outperforms PCA is the work by Belhumeur, Hespanha, and Kriegman \cite{belhumeur1996eigenfaces}. In the context of active learning, this geometric advantage is necessary. Recent evaluations demonstrate that active learning strategies geometrically based on LDA significantly leads to the active learner explicitly focusing on boundary points between classes, maximizing class separation and improving classification performance even when operating on highly imbalanced data with a restricted query budget \cite{tharwat2024using}.

However, as shown here, to fully capture the characteristics of the most effective active learning methods, it is necessary to consider aspects of the data beyond geometric projection. Here we incorporate a quantification of the degree of uncertainty in each unlabeled sample as well. A key method established in the literature is to derive a probability space directly from the predicted class probability vectors of the current model \cite{settles2009active}. The question then becomes how to most effectively and efficiently derive this space during the active learning selection process.

Neural networks (NNs) can be effective for the task of generating the probability space, but they require extensive hyperparameter tuning and robust architectural design to achieve the necessary performance. 
We evaluated various NN architectures, including standard Multi-Layer Perceptrons and basic feedforward networks, but found them inferior to Random Forests for generating a reliable and well-calibrated probability space in this specific active learning setup. While modern Pre-Activation Tabular ResNet architectures eventually produce high-quality probability spaces, they require hundreds of training epochs to reach stability at each active learning query step. This drastically increases the time it takes to complete the active learning cycle, making it prohibitive for deployment in practice.

On the other hand, Random Forests (RF) \cite{breiman2001random} provide an effective, fast, and stable baseline evaluator. Random Forests resist overfitting and do not require many epoch-based training loops, making them ideal for the iterative nature of active learning, especially scenarios where the labeled set grows in small batches. In addition, established research shows that standard random forests can be used in active learning to create a computationally efficient approach for maximizing the joint entropy of a batch of samples while achieving maximal information gain and minimizing information redundancy \cite{nguyen2012efficient}. Because of this, while we test both NNs and RFs to confirm these findings empirically, we emphasize Random Forests as the primary engine for generating the probability space and evaluating dataset accuracy.

Lastly, we test a hybrid approach of integrating predictive entropy with the probability space methodology. For a given sample $x$, the probability space is derived from its predicted class probability vector $\hat{p}(x)$, where the probability for class $c$ over $T$ trees is expressed as $\hat{p}_c(x) = \frac{1}{T} \sum_{t=1}^T I(\hat{y}_t(x) = c)$. Entropy is a classic metric for uncertainty sampling \cite{wu2022entropy}, representing the impurity in the model's prediction. It is computed as a function of this assigned probability vector: $H(x) = - \sum_{i=1}^{C} \hat{p}_i(x) \log \hat{p}_i(x)$, where $C$ is the total number of classes. By weighting the probability space distances by the predictive entropy of the samples, we force the Greedy $K$-center algorithm to prioritize points that are both spatially diverse across the model's learned probability distribution and highly uncertain to the current model. This combined objective balances exploration (diversity via Greedy $K$-center) and exploitation (uncertainty via Entropy).

In Section 2, we formally define the various active learning selection algorithms, specifically looking at how the Greedy $K$-center selection operates across raw feature spaces, LDA-reduced spaces, predictive probability spaces, and an entropy-weighted probability space. In Section 3, we demonstrate why the Random Forest is the optimal baseline we use for this task, followed by an empirical evaluation across synthetic and real-world datasets. Ultimately, our results show that the calculation of diversity within a model-derived predictive probability space, especially when weighted by entropy, consistently outperforms the other options for active learning selection with Greedy $K$-center.
\section{Algorithms}

Our proposed methods modify the space used in the Greedy $K$-center algorithm by transforming the feature vectors into new metric spaces before we calculate diversity. The main focus is to redefine the vectors used in the distance function $d(u, l)$ to better capture the classification utility of a sample. For all versions of this approach, we let $U$ be the pool of available unlabeled instances, $L$ be the currently labeled set, and $x_i$ represent the vector of input features for sample $i$.

In our setup, the Greedy $K$-center algorithm initializes with a randomly selected initial labeled set $L$. For each subsequent query step, we iteratively select a batch of unlabeled points to be labeled and moved into set $L$. To select each point to label, the algorithm greedily chooses the unlabeled instance $u \in U$ that maximizes the minimum distance to the already labeled set $L$:
\begin{equation}
    u^* = \arg\max_{u \in U} \min_{l \in L} d(u, l) \label{eq:kcenter}
\end{equation}
The following subsections detail the different metric spaces used to calculate this distance $d(u,l)$.

\subsection{Random Sampling (Baseline)}
As a baseline for our evaluation, we include standard Random Sampling. Here, the active learning algorithm does not rely on a geometric distance, predictive heuristic, or model uncertainty. Instead, at each query step, an unlabeled instance $u \in U$ is selected uniformly at random to be labeled and added to the set $L$:
\begin{equation}
    u^* \sim \mathcal{U}(U) \label{eq:random}
\end{equation}
Random sampling does not explore the data domain or exploit the model's current state, but it provides us with a unbiased sample of the underlying data distribution. It acts as a performance threshold for the different active learning metric space methods to reliably outperform.

\subsection{Feature Space}
In the baseline \textbf{Feature Space} method, $d(u,l)$ is the $L_2$ Euclidean distance between the standardized raw feature vectors:
\begin{equation}
    d(u, l) = || x_u - x_l ||_2 \label{eq:feat}
\end{equation}
While simple to calculate, the raw distance calculation treats all features with equal importance regardless of their predictive power for the classification variable we are targeting.

\subsection{Linear Discriminant Analysis (LDA) Space}
To address issues with high-dimensional features, a common approach is to perform dimensionality reduction before performing any kind of sampling or clustering algorithms. As we previously discussed, while Principal Component Analysis (PCA) is a standard technique, its unsupervised nature maximizes overall variance across the dataset, which can lead to the destruction of specific decision boundaries required for accurate classification.

Here, we utilize Linear Discriminant Analysis (LDA) to project data into a lower-dimensional latent space. Because LDA is supervised, it searches for a projection matrix $W$ that maximizes the separation between classes while minimizing the variance within each class. This is achieved by maximizing Fisher's criterion, the ratio of the between-class scatter matrix $S_B$ to the within-class scatter matrix $S_W$:
\begin{equation}
    W^* = \arg\max_{W} \frac{|W^T S_B W|}{|W^T S_W W|}
\end{equation}

At each active learning step, a new LDA transformation $f_{LDA}(x) = W^T x$ is trained using only the currently available labeled set $L$. Both the unlabeled pool $U$ and the labeled set $L$ are then transformed into this latent space. The distance metric for Greedy selection then becomes the $L_2$ distance between these linear projections:
\begin{equation}
    d(u, l) = || f_{LDA}(x_u) - f_{LDA}(x_l) ||_2
\end{equation}
While LDA maximizes between-class variance in linearly separable environments, it encounters issues with complex, highly non-linear data distributions. If the classes cannot be separated, the LDA embedding forces overlapping classes into a tight cluster, which hurts the Greedy $K$-center algorithm's ability to find meaningful diverse points.

\subsection{Probability Space}
To improve on the geometric projection methods, we look to established methods of quantifying uncertainty by calculating diversity within the model's current predictive representation of the data.

Let $\hat{p}(x)$ represent the predicted class probability vector for sample $x$ outputted by the base classifier. For a Random Forest, this probability vector is calculated as the mean predicted class probabilities of the trees in the forest. Specifically, if $\hat{y}_t(x)$ is the predicted class output by tree $t$, the probability for class $c$ over $T$ trees is:
\begin{equation}
    \hat{p}_c(x) = \frac{1}{T} \sum_{t=1}^T I(\hat{y}_t(x) = c)
\end{equation}
The distance between two points is calculated as the $L_2$ distance between their resulting probability vectors:
\begin{equation}
    d(u, l) = || \hat{p}(u) - \hat{p}(l) ||_2 \label{eq:prob}
\end{equation}
With this distance function (\ref{eq:prob}) the Greedy algorithm select points that are diverse with respect to the \textit{decision boundary}, rather than only diverse with respect to the feature dimensions. Points that the model confidently predicts as class A will be far away from points confidently predicted as class B, and both of these points will be viewed distinctly from points the model is uncertain about.

\subsection{Probability + Entropy Space}
For our hybrid method, we combine the representation properties of diversity with the boundary focus of uncertainty by weighting the probability distance by the predictive entropy of the unlabeled points. The Shannon entropy $H(u)$ ($C$ represents the total number of classes) provides a measure of the impurity in the model's prediction:
\begin{equation}
    H(u) = - \sum_{i=1}^{C} \hat{p}_i(u) \log \hat{p}_i(u) \label{eq:entropy}
\end{equation}
We modify the Greedy selection criteria to prefer instances that are both far away from existing centers in the probability space \textit{and} have a high uncertainty level. The main goal of this scaling is to shrink the distance metric for points the model is already highly confident about, making them less likely to be chosen:
\begin{equation}
    u^* = \arg\max_{u \in U} \left( H(u) \cdot \min_{l \in L} d(u, l) \right) \label{eq:prob_ent}
\end{equation}

\section{Results}

Our experimental analysis of these methods looks at both synthetic data sets and real-world data sets. In the context of our active learning pipeline, a \textit{query} refers to the algorithmic selection of unlabeled instances to be evaluated, labeled by an oracle, and subsequently added to the training set. A \textit{batch} refers to the specific number of instances selected during a single iterative query step. 

To ensure statistical significance and account for variance, all evaluated metrics are averaged over 100 independent runs. In addition, for each run, the dataset has a different train/test split (always 80/20) and a new random initial labeled pool. We evaluate the proposed methods in two different ways for each dataset: a larger query budget (10\% to 50\% of the training data, adding a batch of 10\% at each query step) and a smaller query budget (100 to 400 total samples, adding a batch of 25 samples at each query step).

\subsection{Baseline Evaluator Selection: RF vs. NN}

Before formally evaluating the various metric spaces, it is necessary for us to establish and justify the base model responsible for generating the predictive probability space and evaluating the queried samples. Table \ref{tab:baselines} details the fully supervised runtime performance of a Random Forest (RF) classifier against a Neural Network (NN) across six distinct datasets. The NN utilizes a modern Pre-Activation Tabular ResNet architecture designed specifically for tabular data, as explained previously.

\begin{table}[htbp]
\caption{Fully Supervised Baselines: Runtime Comparison for RF vs. NN (L4 GPU)}
\centering
\footnotesize
\begin{tabular}{lccc}
\toprule
\textbf{Dataset} & \textbf{Size} & \textbf{RF Time (s)} & \textbf{NN Time (s)} \\
\midrule
Iris & 150 & 0.1617 & 0.2926 \\
Wine & 178 & 0.1657 & 0.6458 \\
Spambase & 4,601 & 0.2362 & 2.6099 \\
Phoneme & 5,404 & 0.2459 & 8.5601 \\
Satimage & 6,435 & 0.2832 & 9.0500 \\
\bottomrule
\end{tabular}
\label{tab:baselines}
\end{table}

As we can see with this table, the Tabular ResNet (NN) requires a high number of training epochs (more than 100 epochs) to achieve basic stability and similar accuracy. In the context of active learning and this Greedy K-center algorithm, where models are repeatedly retrained on small, incrementally growing sets of labeled data, this training requirement creates a severe computational cost. As we can see in Table \ref{tab:baselines}, the NN demands orders of magnitude more computational time to reach convergence. In these evaluations, convergence is measured with early stopping: a 20\% validation subset is monitored during training, and the training loop terminates once the validation loss fails to improve by at least $10^{-4}$ for 20 consecutive epochs, at which point the optimal weights are restored.

Because the Neural Network takes significantly more time to stabilize throughout the active learning training cycle, we will not utilize it to generate the probability space or evaluate the final queries. Instead, the Random Forest demonstrates robust stability in a fraction of the time, which means that we can bypass the need for complex hyperparameter tuning or extended epoch loops. This supports the findings in \cite{nguyen2012efficient} that Random Forests are effective, stable baselines for active learning sampling methodologies. As a result of this exploration, for all subsequent experiments and active learning iterations in this paper, we will utilize the Random Forest both to drive the selection queries via the probability space and to serve as the final baseline evaluator.

\subsection{Large Query Budgets ($K$ Proportional to Dataset Size)}
In active learning, a query budget restricts the total cumulative number of labels the algorithm can request from the oracle. We evaluated the performance of the different metric spaces and methodologies across large active learning query budgets ($K$). In the context of the Greedy $K$-center algorithm, the budget $K$ represents the total size of the labeled set selected by the algorithm. In these experiments, we simulate environments where the total budget $K$ grows from 10\% to 50\% of the total available training data. The algorithm achieves this by selecting a fixed batch size equal to 10\% of the data at each iterative query step.

To test the robustness of these metric spaces, we generated a series of synthetic datasets alongside the real-world sets. Each synthetic dataset consists of 2,500 data points defined across 20 dimensions, with 15 informative features. To evaluate at different complexities of the classification task, we varied the number of Gaussian clusters assigned to each target class, generating datasets with 1, 2, 3, and 4 clusters per class (labeled \textit{Synthetic 1C} through \textit{Synthetic 4C}, respectively). Increasing the clusters per class artificially fragments the underlying data distribution, forcing the decision boundary to become significantly more challenging to map accurately with a limited budget of active learning queries. 

Based on our findings in Section III-A, we configured our active learning pipeline to use Random Forest (RF) as the base model for generating the predictive probabilities, and we evaluated the resulting query batches using a RF to isolate the performance of the metric spaces.

\begin{figure*}[t]
  \centering
  \includegraphics[width=0.48\textwidth]{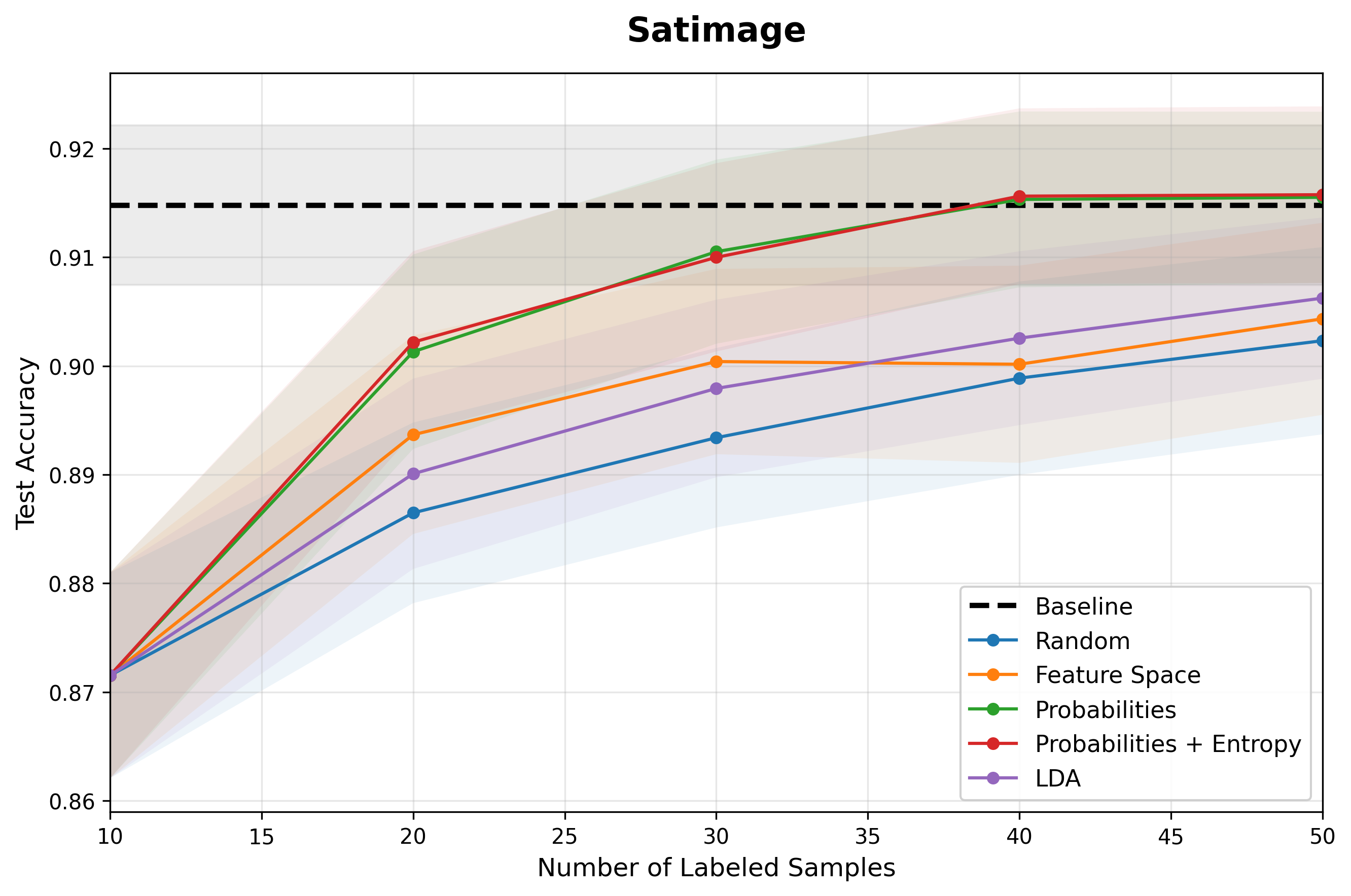}
  \hfill
  \includegraphics[width=0.48\textwidth]{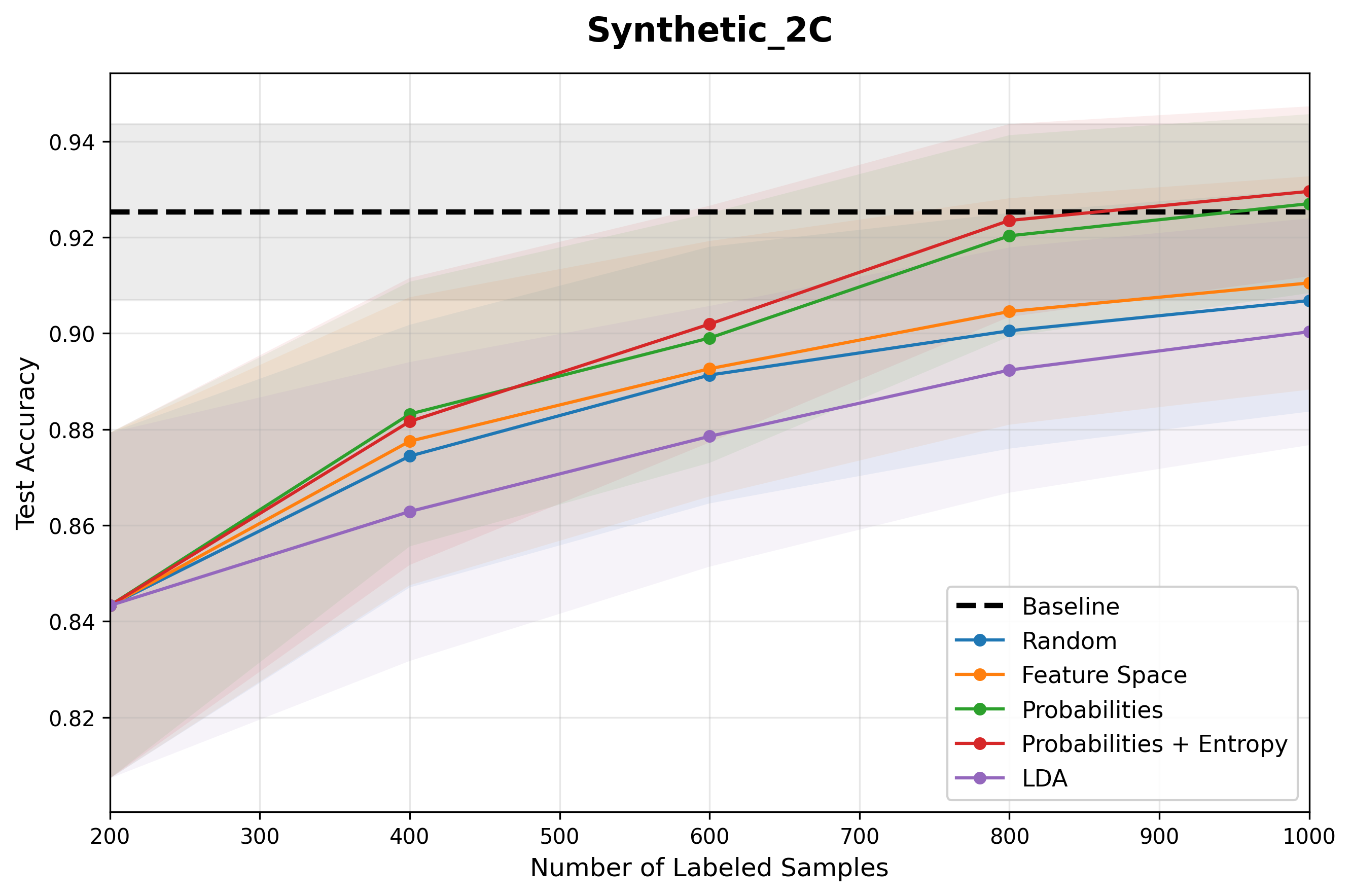}
  \\[0.4cm]
  \includegraphics[width=0.48\textwidth]{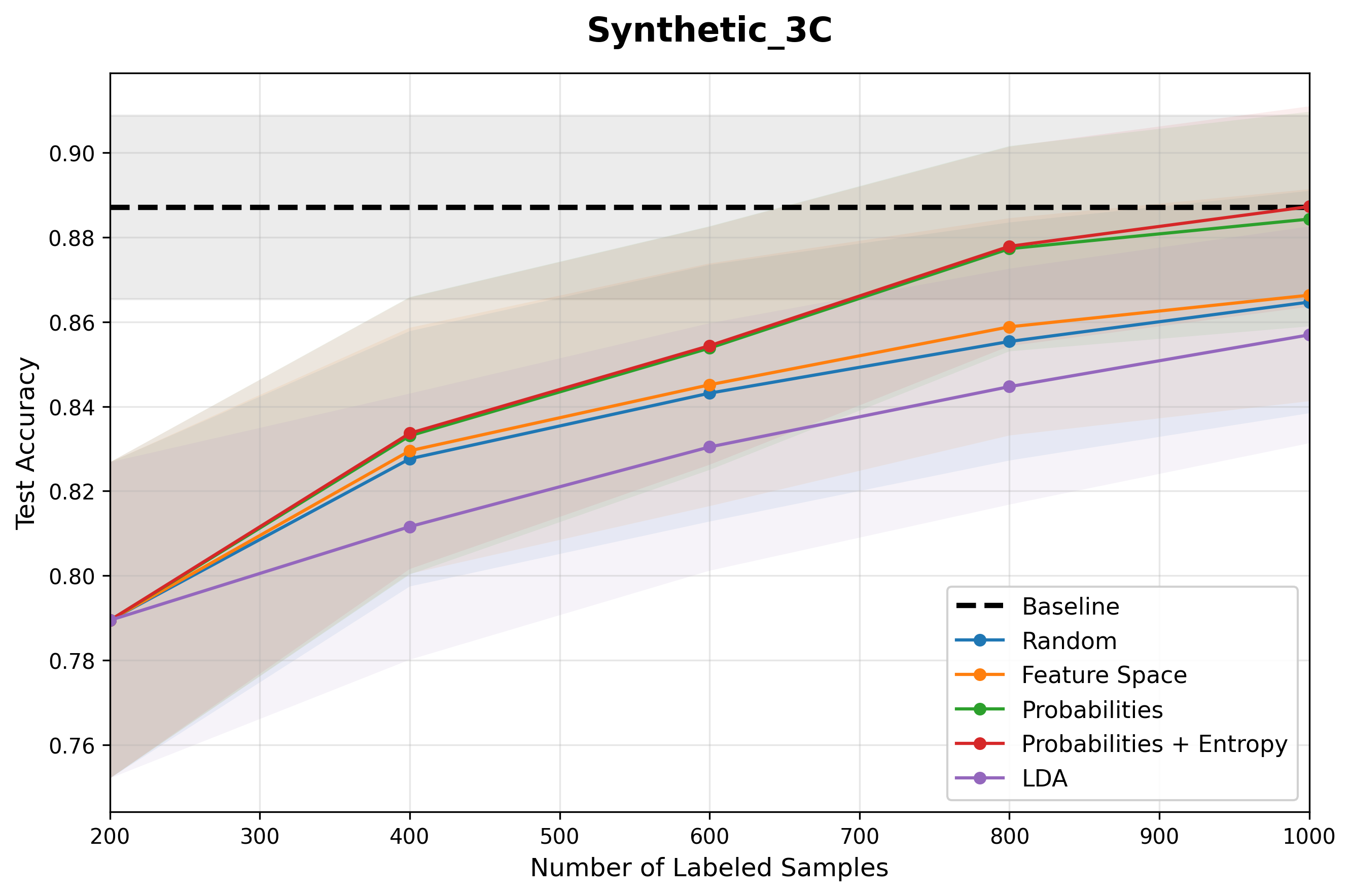}
  \hfill
  \includegraphics[width=0.48\textwidth]{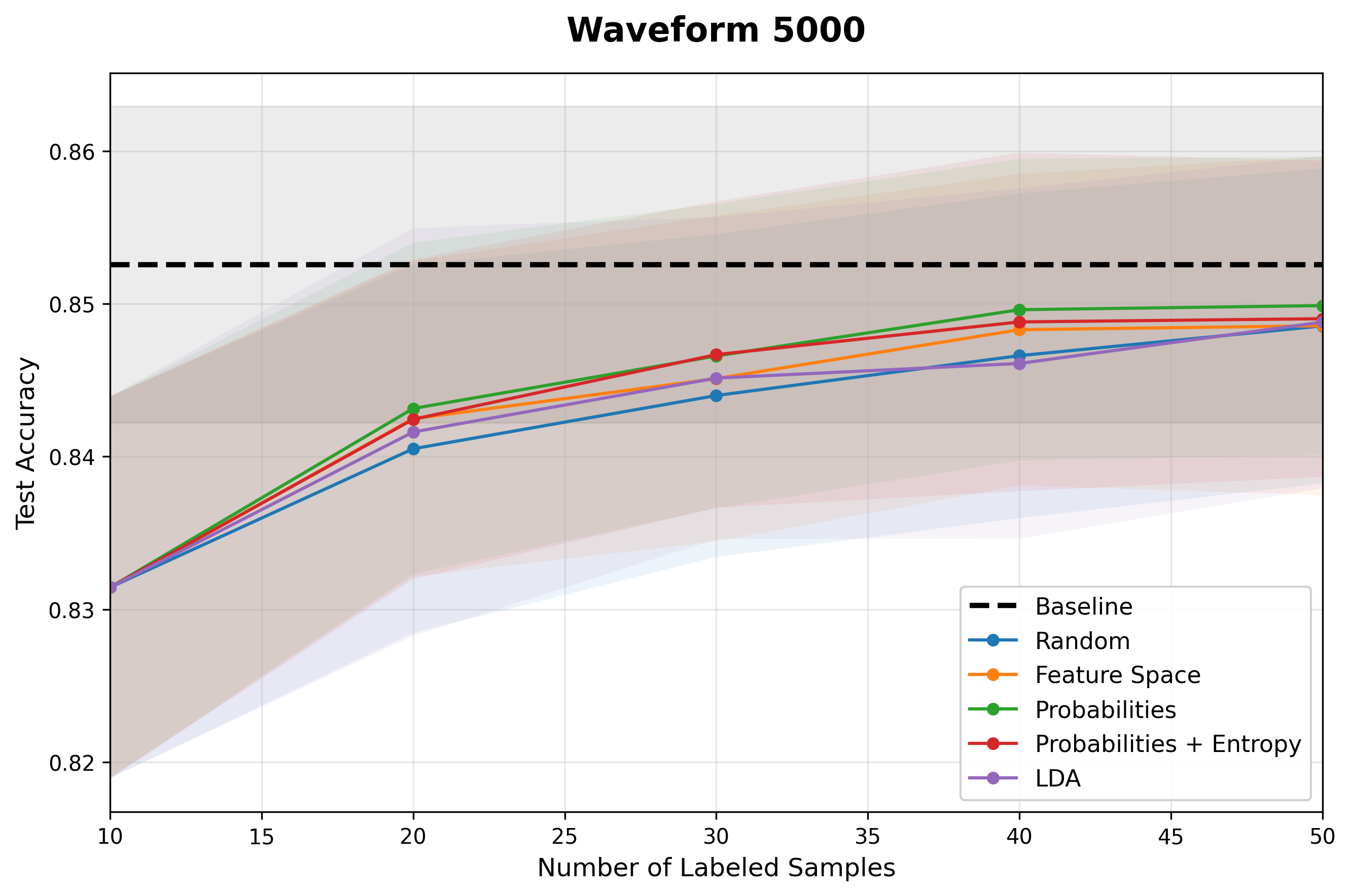}
  \caption{Learning curves for large percentage batches (10\%-50\%). The charts demonstrate the performance advantage of mapping queries into the Probability Space, even as the complexity of the dataset increases (e.g. Synthetic 2C, Synthetic 3C).}
  \label{fig:pct_graphs}
\end{figure*}

The empirical results are detailed in Table \ref{tab:pct_acc} and Table \ref{tab:pct_auc}. These results show us a clear hierarchy in sampling efficacy across the methods. Projecting the unlabeled pool into the Probability Space consistently yields higher terminal accuracy and a higher Area Under the Learning Curve (AUC) compared to simply using the raw Feature Space or supervised LDA projections.

This is also apparent in the learning curves shown in Figure \ref{fig:pct_graphs}. By calculating the Greedy $K$-center distances over class probability vectors instead of geometric coordinates, the active learner focuses on samples that represent diverse, uncertain regions of the actual decision boundary. This bypasses the structural noise and non-informative variance of the physical feature space. As the synthetic datasets increase in complexity (moving from 1C to 4C), the performance gap between the probability spaces and the geometric spaces (Feature Space, LDA) stays large, confirming that raw distance calculations don't do well in comparison, including with non-linear class distributions.

\begin{table*}[t]
\caption{Mean Accuracy at Final Query Step (50\%) (The best results in each row are \textbf{bolded}, and the second-best results are \underline{underlined}).}
\centering
\resizebox{\textwidth}{!}{%
\begin{tabular}{lccccccc}
\toprule
\textbf{Dataset} & \textbf{Size} & \textbf{Baseline} & \textbf{Random} & \textbf{Feature Space} & \textbf{LDA} & \textbf{Prob} & \textbf{Prob+Ent} \\
\midrule
Iris & 150 & 0.951 $\pm$ 0.037 & 0.948 $\pm$ 0.043 & 0.947 $\pm$ 0.038 & 0.942 $\pm$ 0.045 & \textbf{0.951 $\pm$ 0.038} & \underline{0.949 $\pm$ 0.039} \\
Wine & 178 & 0.981 $\pm$ 0.022 & 0.969 $\pm$ 0.032 & 0.966 $\pm$ 0.031 & 0.972 $\pm$ 0.024 & \underline{0.980 $\pm$ 0.023} & \textbf{0.981 $\pm$ 0.022} \\
Digits & 1,797 & 0.973 $\pm$ 0.008 & 0.961 $\pm$ 0.010 & 0.974 $\pm$ 0.007 & 0.974 $\pm$ 0.008 & \textbf{0.978 $\pm$ 0.007} & \underline{0.977 $\pm$ 0.007} \\
Synthetic 1C & 2,500 & 0.969 $\pm$ 0.014 & 0.961 $\pm$ 0.016 & 0.963 $\pm$ 0.015 & 0.960 $\pm$ 0.018 & \textbf{0.972 $\pm$ 0.012} & \underline{0.971 $\pm$ 0.013} \\
Synthetic 2C & 2,500 & 0.925 $\pm$ 0.018 & 0.907 $\pm$ 0.023 & 0.910 $\pm$ 0.022 & 0.900 $\pm$ 0.024 & \underline{0.927 $\pm$ 0.019} & \textbf{0.930 $\pm$ 0.018} \\
Synthetic 3C & 2,500 & 0.887 $\pm$ 0.022 & 0.865 $\pm$ 0.026 & 0.866 $\pm$ 0.025 & 0.857 $\pm$ 0.026 & \underline{0.884 $\pm$ 0.026} & \textbf{0.887 $\pm$ 0.024} \\
Synthetic 4C & 2,500 & 0.862 $\pm$ 0.020 & 0.834 $\pm$ 0.025 & 0.836 $\pm$ 0.023 & 0.823 $\pm$ 0.025 & \underline{0.851 $\pm$ 0.023} & \textbf{0.855 $\pm$ 0.023} \\
Spambase & 4,601 & 0.953 $\pm$ 0.006 & 0.945 $\pm$ 0.007 & 0.944 $\pm$ 0.007 & 0.945 $\pm$ 0.007 & \underline{0.947 $\pm$ 0.007} & \textbf{0.947 $\pm$ 0.007} \\
Waveform-5000 & 5,000 & 0.853 $\pm$ 0.010 & 0.849 $\pm$ 0.010 & 0.849 $\pm$ 0.011 & 0.849 $\pm$ 0.011 & \textbf{0.850 $\pm$ 0.010} & \underline{0.849 $\pm$ 0.010} \\
Phoneme & 5,404 & 0.909 $\pm$ 0.008 & 0.885 $\pm$ 0.009 & \textbf{0.893 $\pm$ 0.008} & 0.886 $\pm$ 0.009 & 0.889 $\pm$ 0.009 & \underline{0.890 $\pm$ 0.009} \\
Satimage & 6,435 & 0.915 $\pm$ 0.007 & 0.902 $\pm$ 0.009 & 0.904 $\pm$ 0.009 & 0.906 $\pm$ 0.007 & \underline{0.916 $\pm$ 0.008} & \textbf{0.916 $\pm$ 0.008} \\
\bottomrule
\end{tabular}%
}
\label{tab:pct_acc}
\end{table*}

\begin{table*}[t]
\caption{Area Under Curve (AUC) for 10\%-50\% Batches (The best results in each row are \textbf{bolded}, and the second-best results are \underline{underlined}).}
\centering
\resizebox{\textwidth}{!}{%
\begin{tabular}{lcccccc}
\toprule
\textbf{Dataset} & \textbf{Size} & \textbf{Random} & \textbf{Feature Space} & \textbf{LDA} & \textbf{Prob} & \textbf{Prob+Ent} \\
\midrule
Iris & 150 & 37.468 $\pm$ 1.502 & \textbf{37.643 $\pm$ 1.507} & 37.352 $\pm$ 1.663 & 37.628 $\pm$ 1.578 & \underline{37.637 $\pm$ 1.571} \\
Wine & 178 & 37.776 $\pm$ 1.273 & 37.679 $\pm$ 1.236 & 37.788 $\pm$ 1.254 & \underline{38.193 $\pm$ 1.063} & \textbf{38.249 $\pm$ 0.994} \\
Digits & 1,797 & 37.597 $\pm$ 0.382 & 38.029 $\pm$ 0.317 & 38.053 $\pm$ 0.322 & \underline{38.398 $\pm$ 0.254} & \textbf{38.462 $\pm$ 0.287} \\
Synthetic 1C & 2,500 & 759.608 $\pm$ 15.231 & 761.548 $\pm$ 14.128 & 758.682 $\pm$ 16.418 & \underline{766.178 $\pm$ 13.350} & \textbf{766.352 $\pm$ 13.667} \\
Synthetic 2C & 2,500 & 708.258 $\pm$ 20.482 & 710.314 $\pm$ 20.896 & 701.114 $\pm$ 21.719 & \underline{717.530 $\pm$ 18.973} & \textbf{718.698 $\pm$ 18.998} \\
Synthetic 3C & 2,500 & 670.652 $\pm$ 22.658 & 672.282 $\pm$ 21.561 & 661.988 $\pm$ 22.685 & \underline{680.222 $\pm$ 21.924} & \textbf{680.862 $\pm$ 21.291} \\
Synthetic 4C & 2,500 & 643.712 $\pm$ 20.974 & 644.942 $\pm$ 19.979 & 634.114 $\pm$ 20.462 & \underline{651.542 $\pm$ 20.162} & \textbf{652.164 $\pm$ 20.087} \\
Spambase & 4,601 & 37.509 $\pm$ 0.295 & 37.450 $\pm$ 0.289 & 37.475 $\pm$ 0.307 & \textbf{37.608 $\pm$ 0.275} & \underline{37.595 $\pm$ 0.277} \\
Waveform-5000 & 5,000 & 33.711 $\pm$ 0.386 & 33.759 $\pm$ 0.369 & 33.730 $\pm$ 0.414 & \textbf{33.800 $\pm$ 0.347} & \underline{33.782 $\pm$ 0.361} \\
Phoneme & 5,404 & 34.566 $\pm$ 0.331 & \textbf{34.860 $\pm$ 0.275} & 34.588 $\pm$ 0.292 & 34.773 $\pm$ 0.300 & \underline{34.782 $\pm$ 0.311} \\
Satimage & 6,435 & 35.657 $\pm$ 0.310 & 35.822 $\pm$ 0.318 & 35.795 $\pm$ 0.303 & \underline{36.207 $\pm$ 0.308} & \textbf{36.214 $\pm$ 0.300} \\
\bottomrule
\end{tabular}%
}
\label{tab:pct_auc}
\end{table*}

Lastly, augmenting the probability space with entropy weighting (\textit{Prob+Ent}) seems to provide a slight, consistent enhancement. By introducing the predictive entropy constraint, the algorithm penalizes the Greedy $K$-center process for selecting instances that the model is already confident about. Instead of looking at queries uniformly across the probability space, the entropy weighting pushes the selection process toward the most ambiguous and informative samples. This accelerates the rate of convergence and gives us higher overall predictive accuracy without sacrificing the boundary-mapping exploration that we get with the $K$-center objective.

\subsection{Small Query Budgets ($K \le 400$)}

To make sure that our findings hold in environments with extreme data scarcity, something that is very valuable to active learning right now, we significantly constrained the total query budget $K$. In these tests, we restricted the algorithm to a maximum budget of $K = 400$ samples. The active learning process initializes with a random pool of 100 samples, and the algorithm selects a small, fixed batch size of 25 samples at each iterative step until the budget $K$ is exhausted. Following the findings established in Section III-A, we used the RF-RF setup for these evaluations.

\begin{figure*}[t]
  \centering
  \includegraphics[width=0.48\textwidth]{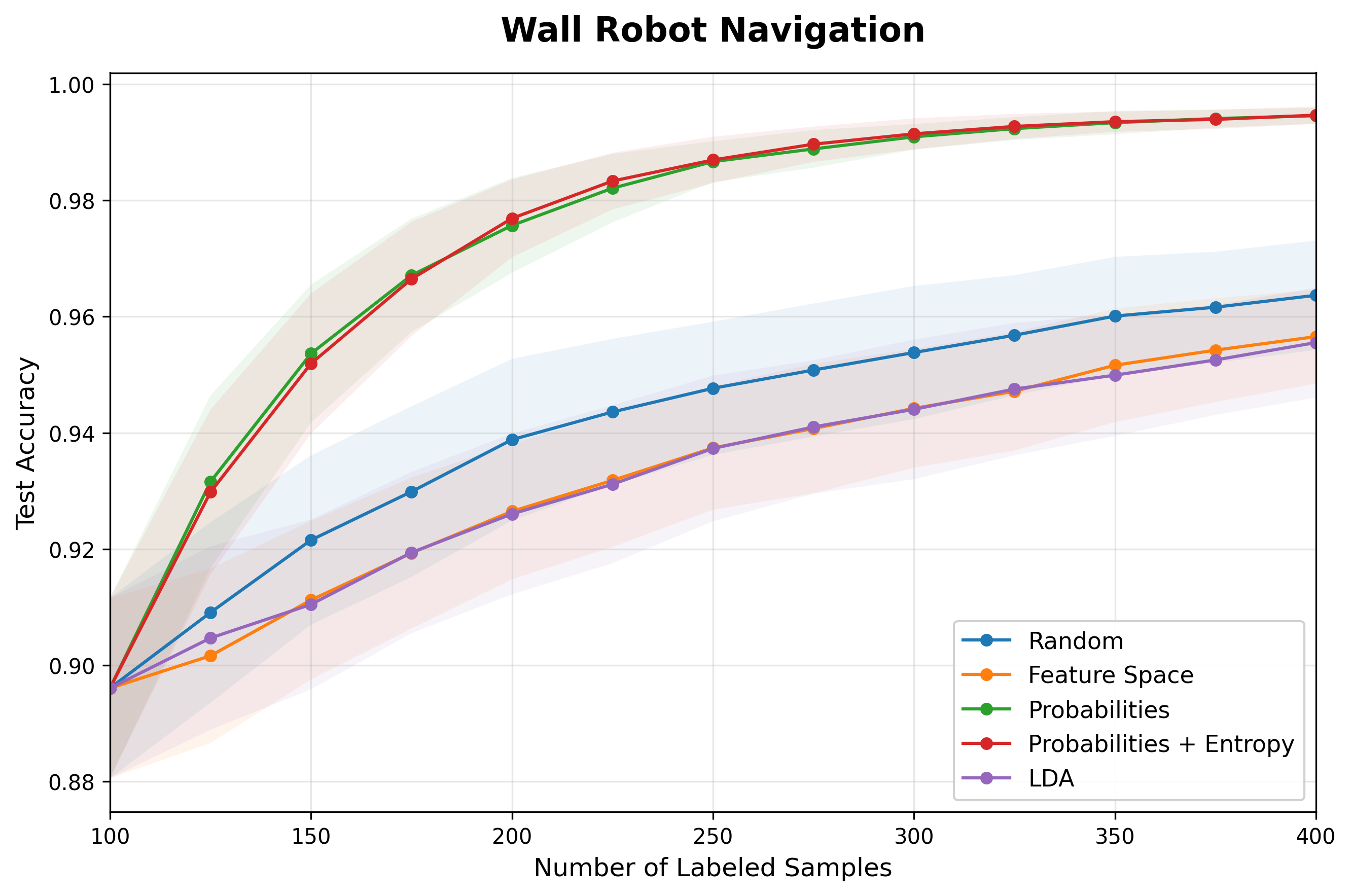}
  \hfill
  \includegraphics[width=0.48\textwidth]{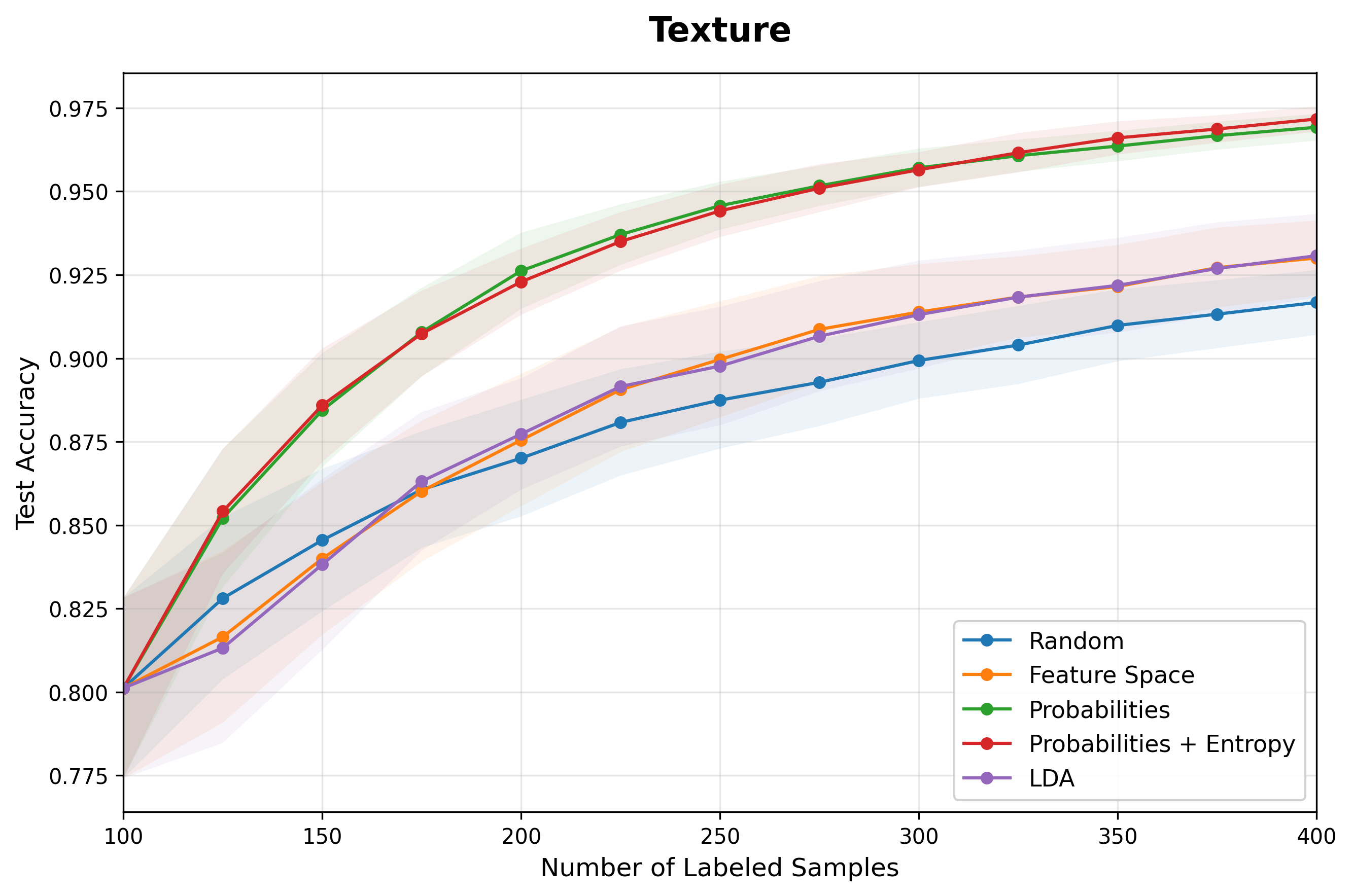}
  \\[0.4cm]
  \includegraphics[width=0.48\textwidth]{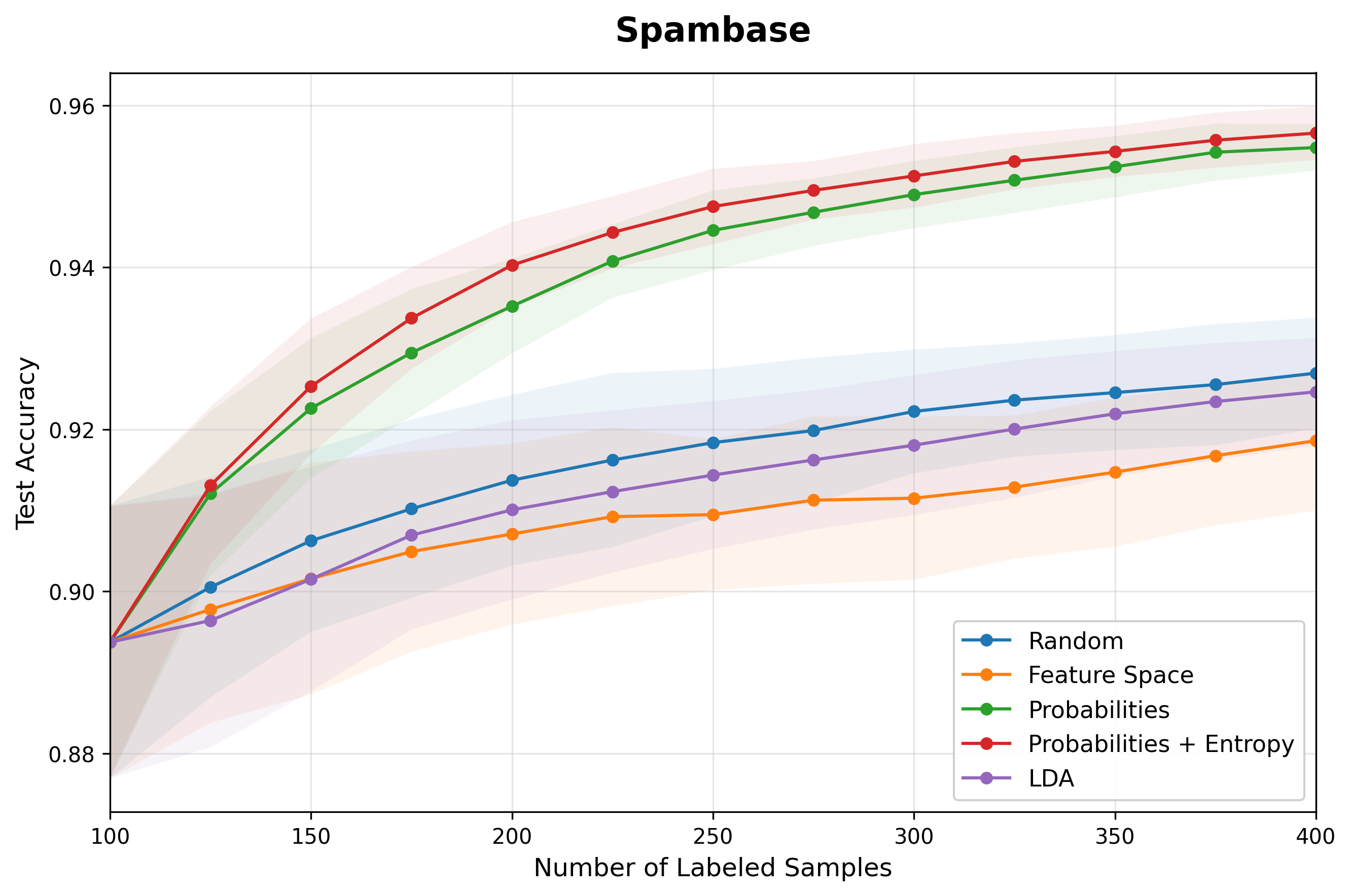}
  \hfill
  \includegraphics[width=0.48\textwidth]{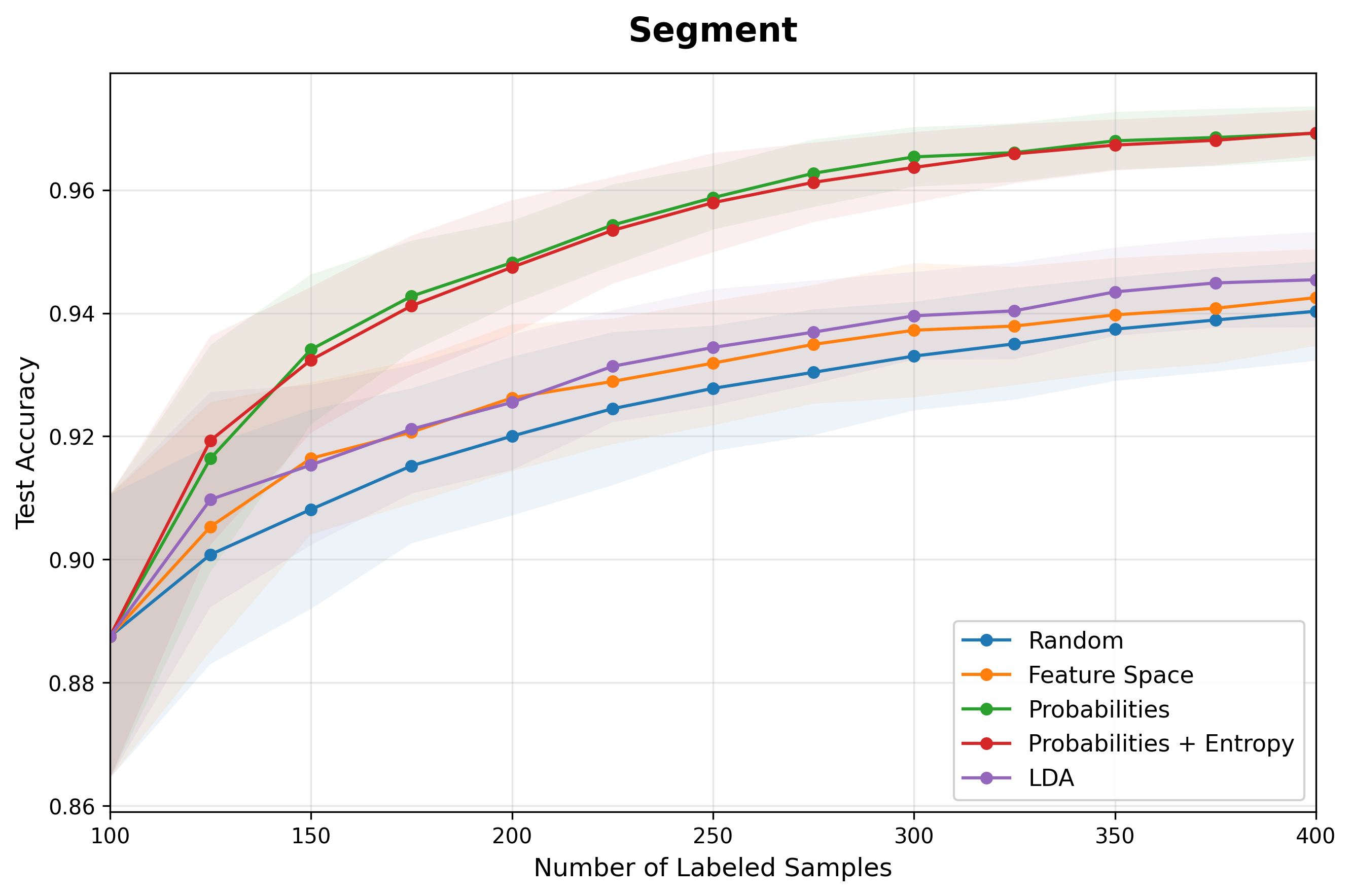}
  \caption{Learning curves (100-400 samples) demonstrating the improvement in performance of using the Probability Space over raw Feature Space and LDA space in the Greedy $K$-center algorithm.}
  \label{fig:scarcity_graphs}
\end{figure*}

The learning curves shown in Figure \ref{fig:scarcity_graphs} show us the impact of metric space selection in these highly constrained environments. Across datasets like \textit{Wall Robot Navigation}, \textit{Texture}, \textit{Spambase}, and \textit{Segment}, mapping the unlabeled pool into the Probability Space gives us a large advantage. For instance, in the \textit{Spambase} dataset, the use of raw Feature Space in Greedy $K$-center actually under performs the purely Random Sampling baseline during early querying phases. This issue occurs because using geometric distance calculations in high-dimensional tabular data can lead to the prioritization of outliers or isolated noisy features that hold no predictive value for the classification task. The Probability Space addresses this vulnerability and drives the active learning model toward optimal test accuracy.

\begin{table*}[t]
\caption{Mean Accuracy at 400 Samples (The best results in each row are \textbf{bolded}, and the second-best results are \underline{underlined}).}
\centering
\begin{tabularx}{\textwidth}{@{} l c *{5}{>{\centering\arraybackslash}X} @{}}
\toprule
\textbf{Dataset} & \textbf{Size} & \textbf{Random} & \textbf{Feature Space} & \textbf{LDA} & \textbf{Prob} & \textbf{Prob+Ent} \\
\midrule
Digits & 1,797 & 0.954 $\pm$ 0.009 & 0.965 $\pm$ 0.007 & 0.971 $\pm$ 0.005 & \underline{0.981 $\pm$ 0.004} & \textbf{0.982 $\pm$ 0.004} \\
Segment & 2,310 & 0.940 $\pm$ 0.008 & 0.943 $\pm$ 0.008 & 0.945 $\pm$ 0.008 & \textbf{0.969 $\pm$ 0.004} & \textbf{0.969 $\pm$ 0.004} \\
1 Cluster Syn & 2,500 & 0.944 $\pm$ 0.022 & 0.948 $\pm$ 0.021 & 0.941 $\pm$ 0.025 & \underline{0.968 $\pm$ 0.017} & \textbf{0.973 $\pm$ 0.014} \\
2 Cluster Syn & 2,500 & 0.875 $\pm$ 0.030 & 0.879 $\pm$ 0.030 & 0.856 $\pm$ 0.034 & \underline{0.900 $\pm$ 0.027} & \textbf{0.909 $\pm$ 0.027} \\
3 Cluster Syn & 2,500 & 0.822 $\pm$ 0.033 & 0.832 $\pm$ 0.032 & 0.803 $\pm$ 0.036 & \underline{0.844 $\pm$ 0.035} & \textbf{0.853 $\pm$ 0.033} \\
4 Cluster Syn & 2,500 & 0.792 $\pm$ 0.027 & 0.792 $\pm$ 0.030 & 0.768 $\pm$ 0.032 & \underline{0.807 $\pm$ 0.031} & \textbf{0.813 $\pm$ 0.033} \\
Madelon & 2,600 & \underline{0.589 $\pm$ 0.025} & 0.588 $\pm$ 0.022 & \textbf{0.590 $\pm$ 0.024} & 0.568 $\pm$ 0.024 & 0.567 $\pm$ 0.020 \\
Spambase & 4,601 & 0.927 $\pm$ 0.007 & 0.919 $\pm$ 0.009 & 0.925 $\pm$ 0.007 & \underline{0.955 $\pm$ 0.003} & \textbf{0.957 $\pm$ 0.003} \\
Waveform-5000 & 5,000 & 0.841 $\pm$ 0.008 & 0.840 $\pm$ 0.008 & 0.838 $\pm$ 0.009 & \textbf{0.847 $\pm$ 0.007} & \underline{0.847 $\pm$ 0.009} \\
Phoneme & 5,404 & 0.829 $\pm$ 0.010 & 0.831 $\pm$ 0.012 & 0.825 $\pm$ 0.010 & \underline{0.858 $\pm$ 0.008} & \textbf{0.860 $\pm$ 0.008} \\
Wall-Robot-Nav & 5,456 & 0.964 $\pm$ 0.009 & 0.957 $\pm$ 0.008 & 0.956 $\pm$ 0.010 & \textbf{0.995 $\pm$ 0.001} & \underline{0.995 $\pm$ 0.002} \\
Page-blocks & 5,473 & 0.959 $\pm$ 0.005 & 0.971 $\pm$ 0.003 & 0.973 $\pm$ 0.003 & \textbf{0.977 $\pm$ 0.002} & \textbf{0.977 $\pm$ 0.002} \\
Texture & 5,500 & 0.917 $\pm$ 0.010 & 0.930 $\pm$ 0.011 & 0.931 $\pm$ 0.013 & \underline{0.969 $\pm$ 0.004} & \textbf{0.972 $\pm$ 0.004} \\
Optdigits & 5,620 & 0.932 $\pm$ 0.009 & 0.941 $\pm$ 0.009 & 0.952 $\pm$ 0.006 & \underline{0.964 $\pm$ 0.004} & \textbf{0.968 $\pm$ 0.004} \\
Satimage & 6,435 & 0.877 $\pm$ 0.006 & 0.869 $\pm$ 0.013 & 0.875 $\pm$ 0.011 & \textbf{0.901 $\pm$ 0.005} & \underline{0.894 $\pm$ 0.009} \\
Ringnorm & 7,400 & 0.931 $\pm$ 0.009 & 0.735 $\pm$ 0.041 & 0.862 $\pm$ 0.022 & \underline{0.955 $\pm$ 0.004} & \textbf{0.958 $\pm$ 0.004} \\
Twonorm & 7,400 & 0.960 $\pm$ 0.005 & 0.960 $\pm$ 0.004 & 0.959 $\pm$ 0.005 & \textbf{0.966 $\pm$ 0.003} & \textbf{0.966 $\pm$ 0.003} \\
Pendigits & 10,992 & 0.943 $\pm$ 0.007 & 0.939 $\pm$ 0.009 & 0.949 $\pm$ 0.007 & \textbf{0.979 $\pm$ 0.004} & \underline{0.972 $\pm$ 0.004} \\
EEG-Eye-State & 14,980 & 0.749 $\pm$ 0.011 & 0.708 $\pm$ 0.016 & 0.699 $\pm$ 0.017 & \underline{0.767 $\pm$ 0.008} & \textbf{0.775 $\pm$ 0.008} \\
Magic Telescope & 19,020 & 0.835 $\pm$ 0.008 & 0.805 $\pm$ 0.015 & 0.826 $\pm$ 0.012 & \underline{0.851 $\pm$ 0.004} & \textbf{0.851 $\pm$ 0.003} \\
\bottomrule
\end{tabularx}
\label{tab:scar_acc}
\end{table*}

\begin{table*}[t]
\caption{Area Under Curve (AUC) for 100-400 Samples (The best results in each row are \textbf{bolded}, and the second-best results are \underline{underlined}).}
\centering
\begin{tabularx}{\textwidth}{@{} l c *{5}{>{\centering\arraybackslash}X} @{}}
\toprule
\textbf{Dataset} & \textbf{Size} & \textbf{Random} & \textbf{Feature Space} & \textbf{LDA} & \textbf{Prob} & \textbf{Prob+Ent} \\
\midrule
Digits & 1,797 & 278.251 $\pm$ 3.268 & 279.356 $\pm$ 2.112 & 281.919 $\pm$ 2.162 & \underline{286.673 $\pm$ 1.407} & \textbf{287.086 $\pm$ 1.400} \\
Segment & 2,310 & 277.127 $\pm$ 2.608 & 278.377 $\pm$ 2.534 & 278.982 $\pm$ 2.003 & \textbf{285.340 $\pm$ 1.372} & \underline{285.158 $\pm$ 1.632} \\
1 Cluster Syn & 2,500 & 278.283 $\pm$ 7.539 & 279.770 $\pm$ 6.906 & 277.384 $\pm$ 8.824 & \underline{284.644 $\pm$ 6.711} & \textbf{286.439 $\pm$ 6.102} \\
2 Cluster Syn & 2,500 & 254.503 $\pm$ 9.648 & 255.605 $\pm$ 9.808 & 248.800 $\pm$ 11.208 & \underline{259.561 $\pm$ 9.396} & \textbf{261.957 $\pm$ 9.252} \\
3 Cluster Syn & 2,500 & 238.549 $\pm$ 10.123 & 240.371 $\pm$ 10.122 & 233.694 $\pm$ 10.800 & \underline{242.694 $\pm$ 10.739} & \textbf{244.407 $\pm$ 10.680} \\
4 Cluster Syn & 2,500 & 227.725 $\pm$ 9.596 & 228.457 $\pm$ 9.671 & 222.446 $\pm$ 9.735 & \underline{230.420 $\pm$ 9.664} & \textbf{232.262 $\pm$ 9.299} \\
Madelon & 2,600 & 167.707 $\pm$ 5.650 & \underline{167.749 $\pm$ 5.378} & \textbf{168.196 $\pm$ 5.870} & 163.971 $\pm$ 5.137 & 164.051 $\pm$ 5.120 \\
Spambase & 4,601 & 274.785 $\pm$ 2.289 & 272.584 $\pm$ 2.837 & 273.762 $\pm$ 2.422 & \underline{281.555 $\pm$ 0.987} & \textbf{282.333 $\pm$ 0.911} \\
Waveform-5000 & 5,000 & 248.768 $\pm$ 2.335 & 248.709 $\pm$ 2.251 & 247.560 $\pm$ 2.691 & \textbf{251.212 $\pm$ 1.782} & \underline{251.165 $\pm$ 1.913} \\
Phoneme & 5,404 & 243.345 $\pm$ 2.973 & 242.699 $\pm$ 3.032 & 242.614 $\pm$ 2.712 & \underline{249.234 $\pm$ 2.044} & \textbf{250.104 $\pm$ 2.206} \\
Wall-Robot-Nav & 5,456 & 282.590 $\pm$ 2.739 & 279.804 $\pm$ 2.441 & 279.749 $\pm$ 2.910 & \underline{292.543 $\pm$ 1.300} & \textbf{292.552 $\pm$ 1.147} \\
Page-blocks & 5,473 & 285.291 $\pm$ 1.696 & 288.668 $\pm$ 1.310 & 289.333 $\pm$ 1.162 & \textbf{290.662 $\pm$ 0.838} & \underline{290.473 $\pm$ 0.910} \\
Texture & 5,500 & 263.766 $\pm$ 3.616 & 265.939 $\pm$ 4.149 & 265.844 $\pm$ 4.205 & \underline{278.460 $\pm$ 2.009} & \textbf{278.495 $\pm$ 1.968} \\
Optdigits & 5,620 & 272.245 $\pm$ 2.544 & 272.365 $\pm$ 3.192 & 275.472 $\pm$ 2.533 & \underline{280.095 $\pm$ 1.508} & \textbf{281.194 $\pm$ 1.546} \\
Satimage & 6,435 & 258.598 $\pm$ 2.121 & 258.910 $\pm$ 3.094 & 258.368 $\pm$ 2.881 & \textbf{265.025 $\pm$ 1.432} & \underline{263.596 $\pm$ 2.165} \\
Ringnorm & 7,400 & 275.320 $\pm$ 2.267 & 245.611 $\pm$ 9.150 & 257.871 $\pm$ 6.832 & \underline{281.023 $\pm$ 1.278} & \textbf{282.461 $\pm$ 0.941} \\
Twonorm & 7,400 & 285.788 $\pm$ 1.446 & 285.828 $\pm$ 1.171 & 285.537 $\pm$ 1.450 & \underline{287.631 $\pm$ 0.853} & \textbf{288.294 $\pm$ 0.546} \\
Pendigits & 10,992 & 274.106 $\pm$ 2.733 & 274.147 $\pm$ 2.596 & 276.017 $\pm$ 2.650 & \textbf{283.453 $\pm$ 1.961} & \underline{282.363 $\pm$ 1.918} \\
EEG-Eye-State & 14,980 & 212.981 $\pm$ 3.340 & 205.588 $\pm$ 4.465 & 202.501 $\pm$ 4.568 & \underline{216.452 $\pm$ 2.868} & \textbf{217.985 $\pm$ 2.990} \\
Magic Telescope & 19,020 & 246.211 $\pm$ 2.538 & 239.649 $\pm$ 4.213 & 243.560 $\pm$ 3.540 & \underline{250.169 $\pm$ 1.336} & \textbf{250.522 $\pm$ 1.170} \\
\bottomrule
\end{tabularx}
\label{tab:scar_auc}
\end{table*}

Tables \ref{tab:scar_acc} and \ref{tab:scar_auc} confirm these trends. The Probability Space and Probability + Entropy methods consistently outperform the Random, Feature Space, and LDA baselines across both the final accuracy at the 400 sample threshold and the overall Area Under the Curve (AUC). We can see that the \textit{Prob+Ent} methodology frequently captures the absolute highest performance rank. By restricting the Greedy $K$-center algorithm from redundantly querying high-confidence, non-informative samples, the entropy weighting seems to again provide a consistent boost to early-stage active learning without introducing any significant computational overhead.

\subsection{Anomalies}

While the Probability Space methodology demonstrated widespread dominance, there were two notable exceptions across our tests.

First, in the large-batch runs (Section III-B), the \textit{Phoneme} dataset had a deviation from the overall trend. As shown in Table \ref{tab:pct_acc}, selecting queries directly from the raw Feature Space achieved the highest final accuracy ($0.893 \pm 0.008$), barely outperforming the Probability Space ($0.889 \pm 0.009$). The \textit{Phoneme} dataset is highly dense but also low-dimensional, containing only 5 continuous physical attributes. In such these tightly constrained and low-dimensional environments, the physical Euclidean distance between raw features can be highly correlated with class boundaries. As a result, forcing a transformation into a secondary probability space might not help, but we see pretty similar results from both spaces for this dataset.

\begin{figure}[htbp]
  \centering
  \includegraphics[width=\columnwidth]{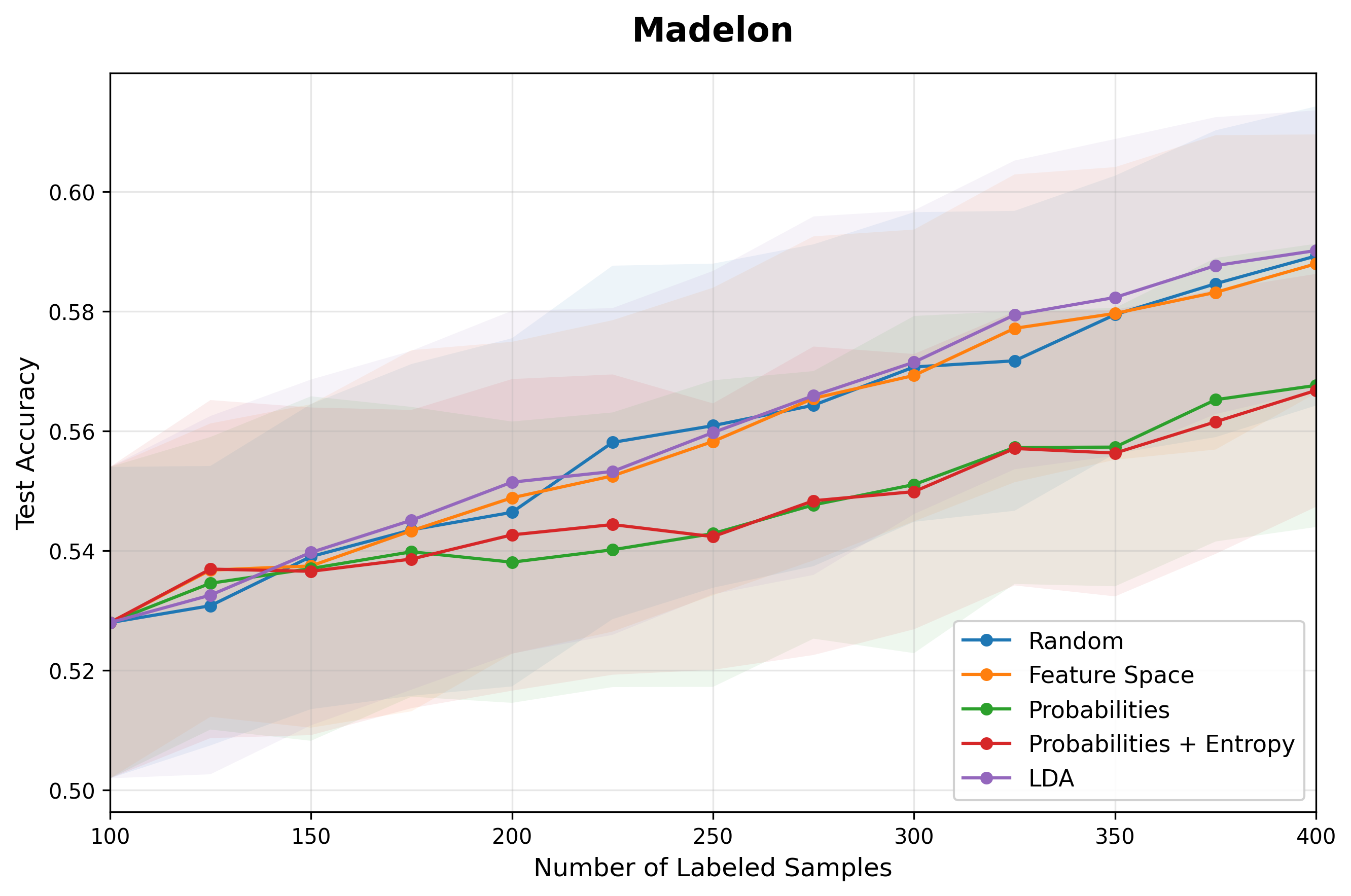}
  \caption{The Madelon dataset, an exception to the Probability Space dominance in extreme data scarcity.}
  \label{fig:madelon_curves}
\end{figure}

Second, in the data scarcity evaluations (Section III-C), the \textit{Madelon} dataset was the one test where the Probability Space underperformed all the other methods, as shown in Figure \ref{fig:madelon_curves}. Madelon is a complex artificial dataset containing data points grouped in 32 clusters placed on the vertices of a five-dimensional hypercube, with heavy added noise. Random Forests require sufficient depth and tree variance to resolve this complex topology. At extremely low amounts of data, the RF model is essentially predicting pure noise. Because of this, the Probability Space transformation is simply mapping noise vectors to noise vectors, which causes the $K$-center Greedy algorithm to select mostly uninformative points. We see that the Probability space methods require the base model to have at least a foundational understanding of the data to function effectively, otherwise it could just amplify the model's initial confusion.

\section{Conclusion}

The objective of active learning is to minimize annotation costs by strategically querying the most informative samples. While diversity-based selection via the Greedy $K$-center algorithm provides a good approximation for optimal subset coverage, our evaluation has shown that the level of success of this algorithm is determined by the metric space in which it operates. 

Our empirical results, tested extensively across both broad percentage-based large batch experiments and extreme data scarcity, establish that transforming the unlabeled pool into a predictive probability space allows the Greedy $K$-center algorithm to bypass the geometric noise inherent to raw features. This model-derived space changes the diversity search to prioritize samples that map distinct, meaningful regions of the classification decision boundary. Across nearly all datasets evaluated, the probability space methodology resulted in substantially better and accelerated active learning convergence. The one exception, the \textit{Madelon} dataset, also seems to reinforce this. In this dataset, at extremely low query budgets, the Random Forest's predictive accuracy is near random. This indicates that the model has failed to extract an initial predictive signal and the probability space just maps pure noise, which means that it requires at least a foundational classification understanding to be effective.

Ultimately, we arrive at the conclusion that the predictive probability space is the most effective metric space for active learning selection when paired with a robust baseline evaluator like the Random Forest. In addition, within this probability space, we found that augmenting the distance calculations with Shannon entropy weighting leads to a very advantageous hybrid balance of exploration and exploitation. Weighting by entropy restricts the Greedy $K$-center algorithm from redundantly querying instances that the model is already confident about. With the integration of this entropy weighting, we see that in the worst-case scenarios, the model is no worse off than using the probability space with no entropy weighting, and in the majority of the datasets, the entropy penalty explicitly yields slightly better \& faster convergence results. When using the Greedy $K$-center algorithm, active learning frameworks relying on diversity constraints should measure distances within these model-interpreted probability boundaries rather than coordinates from the feature space.

\section*{Acknowledgment}

The second author was supported in part by AI institute NSF award 2112533.

\end{document}